\documentclass[conference]{IEEEtran}

\usepackage{cite}
\usepackage{amsmath,amssymb,amsfonts}
\usepackage{amsthm}
\usepackage{graphicx}
\usepackage{textcomp}
\usepackage{xcolor}
\usepackage{subcaption}
\usepackage{booktabs}

\def\BibTeX{{\rm B\kern-.05em{\sc i\kern-.025em b}\kern-.08em
    T\kern-.1667em\lower.7ex\hbox{E}\kern-.125emX}}

\usepackage[linesnumbered,ruled]{algorithm2e}
\DeclareMathOperator*{\argmax}{arg\,max}
\usepackage{url}
\usepackage{multirow}
\usepackage[export]{adjustbox}
\newtheorem{definition}{Definition}
\theoremstyle{definition}

\begin{document}

\title{Active Learning with Combinatorial Coverage}

\author{\IEEEauthorblockN{Sai Prathyush Katragadda}
\IEEEauthorblockA{\textit{Grado Department of Industrial and Systems Engineering} \\
\textit{Virginia Tech}\\
Blacksburg Virginia, USA \\
}
\and
\IEEEauthorblockN{Tyler Cody, Peter Beling, Laura Freeman}
\IEEEauthorblockA{\textit{Virginia Tech National Security Institute} \\
\textit{Virginia Tech}\\
Arlington Virginia, USA \\
}
}

\maketitle

\begin{abstract}
Active learning is a practical field of machine learning that automates the process of selecting which data to label.  Current methods are effective in reducing the burden of data labeling but are heavily model-reliant. This has led to the inability of sampled data to be transferred to new models as well as issues with sampling bias. Both issues are of crucial concern in machine learning deployment. We propose active learning methods utilizing combinatorial coverage to overcome these issues. The proposed methods are data-centric, as opposed to model-centric, and through our experiments we show that the inclusion of coverage in active learning leads to sampling data that tends to be the best in transferring to better performing models and has a competitive sampling bias compared to benchmark methods.

\end{abstract}


\section{Introduction}

Data preprocessing, which involves data labeling, is often the most expensive aspect of deploying machine learning\cite{polyzotis2018data}. Active learning is a sub-field of machine learning that proposes a solution to this problem; it is concerned with selecting which unlabeled datapoints would be most beneficial to label and train on\cite{settles1995active}. This is especially applicable when there is a large pool of unlabeled data, but there is a restriction on the number of datapoints that can be labeled due to budget or time constraints. 

Active learning has found success in many applications including image segmentation \cite{yang2017suggestive}, sequence labeling \cite{settles2008analysis}, medical image classification \cite{hoi2006batch}, cybersecurity \cite{zhao2013cost, nissim2015boosting}, and manufacturing \cite{dasari2021active}. Yet, active learning methods are heavily model-dependent, thus datapoints labeled for one model may not be effective for the training of other models \cite{lowell2019practical, pmlr-v16-tomanek11a}. As pointed out by Paleyes and Urma, model selection is often an iterative process in machine learning deployment \cite{paleyes2020challenges}. Therefore, for the use of active learning to be practical, the sampled datapoints should also be effective in training other models. 

Current active learning methods are successful in particular applications, in other words, a specific method will fare well for some combination of dataset and model \cite{lowell2019practical,pmlr-v16-tomanek11a}. The issue is that, given a dataset, identifying which active learning method is the most effective for a given model may defeat the purpose of applying active learning due to the required investment of resources. This is especially true in deployment scenarios where model type is being constantly updated, e.g., from decision tree to random forest to support vector machine, and so on. 

The cause of this issue is the model dependency of active learning methods. Typically, points to be labeled are regarded as beneficial to a specific model. However, these labeled points may not be the ideal datapoints for training a different model. Additionally, these points may be from a particular area of the feature space, creating sampling bias. Several methods have been proposed in the literature to combat the sampling bias issue, few of which are generalizable to any model and none of which are model independent \cite{settles2008analysis,dasgupta2008hierarchical,krishnan2021mitigating,sener2018active,agarwal2020contextual,Liu_2021_ICCV,Elhamifar_2013_ICCV}. To our knowledge, no data-centric active learning methods have been proposed to sample data so that other models are applicable without resampling.

In short, active learning methods optimize data labeling for a given model, but struggle to sample data which is also effective for training other models. This issue is closely related to the sampling bias issue, which results from the model dependency of existing active learning methods. In this work we propose an active learning approach based on combinatorial coverage (CC) that is data-centric, can generalize to any model, samples data which can be effectively transferred to new models, and achieves improvements in sampling bias. We contribute three CC-based active learning methods:  
\begin{itemize}
\item coverage density sampling,
\item informative coverage density sampling, and
\item uncertainty sampling weighted by coverage density.
\end{itemize} 
While combinatorial interaction testing (CIT) and CC are not widespread in machine learning, several applications have proven successful \cite{pei2017deepxplore, ma2018deepgauge, ma2019deepct, kuhn2020combinatorial,lanus2021combinatorial, cody2022systematic}. We leverage these ideas to develop the proposed methods, and present their competitive performance in terms of classification performance of the trained model and different models as well as the advantages in sampling bias.

The rest of this paper is organized as follows. Next, background is given on active learning and CIT. Then, three CC-based active learning methods are proposed. Subsequently, the experimental design is described and results are presented. The results cover 6 publicly available data sets. Then, before concluding with a synopsis, results and future work are discussed.


\section{Background}

\subsection{Active Learning}
\begin{figure}[t]
    \centering
    \includegraphics[scale=0.6]{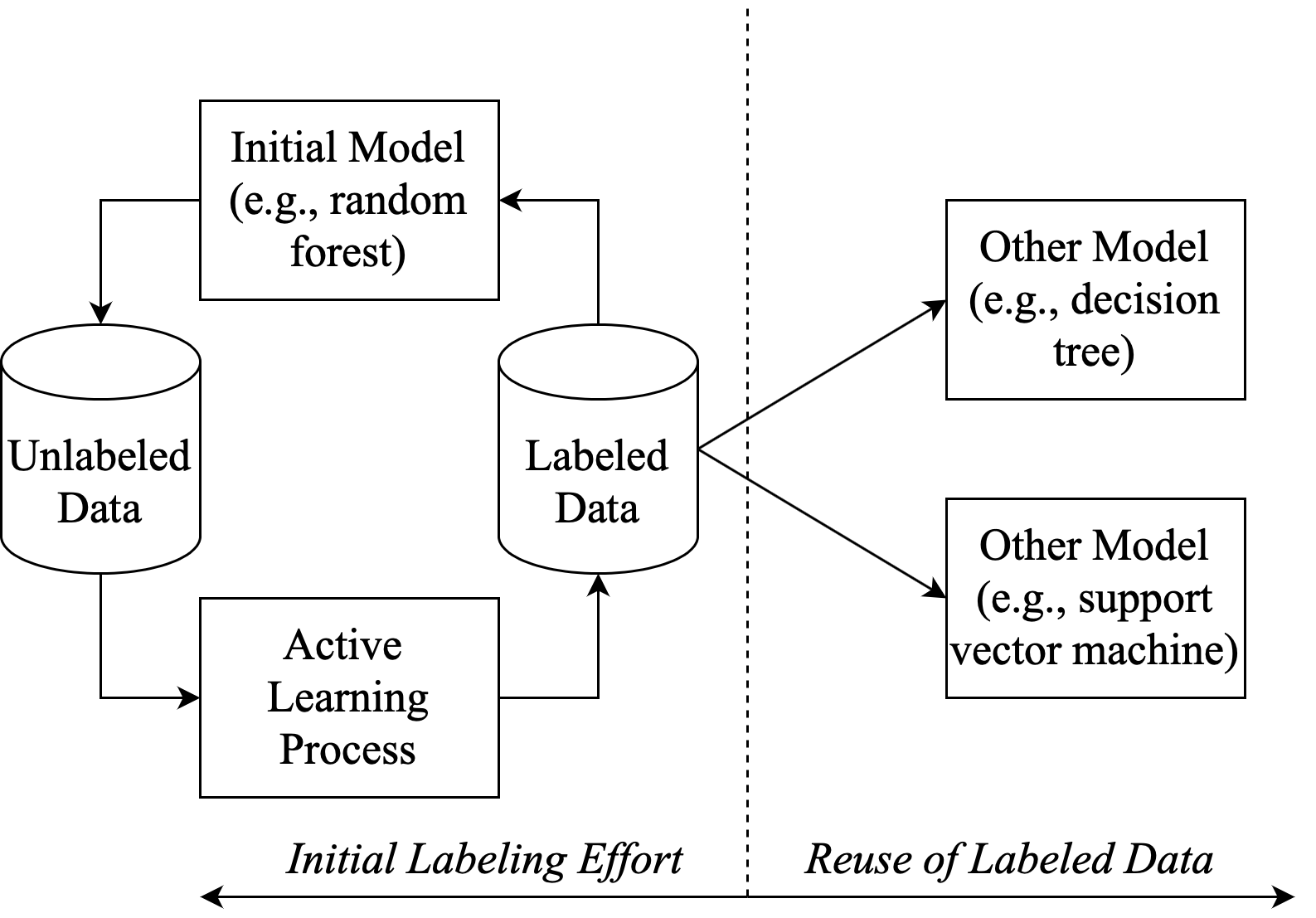}
    \caption{A depiction of active learning for labeling data and the reuse of those labels.}
    \label{fig:AL_wire}
\end{figure}

Active learning methods can be divided into three groups; membership query synthesis, stream based selective sampling, and pool based sampling \cite{settles1995active}. In membership query synthesis the model can arbitrarily select datapoints to label. Stream based selective sampling involves the model receiving a stream of datapoints and deciding whether they should be labeled one at a time. Pool based sampling, which is the focus of this work, involves the model drawing a set of unlabeled samples to label from the entire pool of unlabeled samples available. Several popular and generalizable query strategies exist for active learning. Uncertainty sampling selects the data point the model is currently most uncertain about \cite{lewis1994heterogeneous}. Query by committee selects points to label as those which a committee of classifiers most disagree on or are on average most uncertain about \cite{seung1992query}. 

The dependency of active learning on the model being used has lead to issues in data transferability and sampling bias. The process of sampling data with respect to one model and using it for other models is shown in Figure \ref{fig:AL_wire}, where data is sampled according to the initial model and that same pool of labeled data is used to train different models downstream in the model deployment life cycle. Solutions to transferability have not been proposed but there has been work done to look into the specifics of how well the data selected by popular methods transfers to other models \cite{lowell2019practical, baldridge2004active, pmlr-v16-tomanek11a, pardakhti2021practical}. Lowell et al. find that, in generic classification problems, transferability of data sampled using uncertainty sampling is not guaranteed \cite{lowell2019practical}. Baldridge and Miles share a similar finding; data generated by random sampling are more transferrable to other models than data generated by uncertainty sampling \cite{baldridge2004active}. Tomanek and Morik draw a similar conclusion, but they find that data sampled using uncertainty sampling is transferable to other models for some tasks \cite{pmlr-v16-tomanek11a}. Pardakhti et al. also reference the inability of active learning methods to sample data that is effective for multiple models, but their work focuses on finding the optimal hyper parameters for a model given a data set and active learning method \cite{pardakhti2021practical}. 

The dependency of active learning on the model being used also leads to issues with sampling bias. Several methods have been proposed in the literature to counteract this. To improve the sampling bias of any baseline sampling method Settles and Craven propose an information density method which is computationally expensive for large pools of unlabeled data \cite{settles2008analysis}. Several other successful methods have been proposed to combat the sampling bias and robustness issues, but all these methods are designed to work with specific model types, e.g., with convolutional neural networks \cite{dasgupta2008hierarchical,krishnan2021mitigating,sener2017active,agarwal2020contextual,Liu_2021_ICCV}. A generalizable method for reducing sampling bias is proposed by Elhamifar et al. \cite{Elhamifar_2013_ICCV}, but it involves an optimization problem with $n^2$ variables where $n$ is the number of datapoints in the data set, so the method is not scalable to large data sets.

In this manuscript, the proposed CC-based active learning algorithms are compared to random sampling, uncertainty sampling \cite{settles1995active}, query by committee \cite{seung1992query}, and information density \cite{settles2008analysis}. For pool based active learning, all methods must select some subset of datapoints from an (unlabeled) query set of data. We implement them as follows.
\begin{itemize}
    \item For \emph{random sampling}, we assume a uniform distribution over all datapoints and select datapoints from the query set with equal probability.
    \item For \emph{uncertainty sampling}, we use entropy, defined as $-\sum_{y \in Y}p(y)log(p(y))$ where $y$ is a class and $Y$ is all classes. The model trained on currently labeled data is tested on the query set to determine the probability of each query datapoint belonging to each class. These probabilities are then used for the entropy calculation, and those datapoints for which the model has the highest entropy are selected from the query set. 
    \item For \emph{query by committee}, we also use entropy. Entropy is used to determine which datapoint the committee of classifiers is, on average, most uncertain about. The committee is comprised of three classifiers: random forest, k-nearest neighbors, and logistic regression. The datapoints with the highest average entropy are selected from the query set.
    \item For \emph{information density}, as presented by Settles and Craven \cite{settles2008analysis}, we weight an informativeness measure by a similarity metric. We weight the entropy for a datapoint with its cosine similarity from the labeled data divided by the cardinality of the unlabeled set.
\end{itemize}
We compare these benchmark methods to the proposed methods using 6 open-source data sets that are described in Section \ref{sec:ed}.

\subsection{Combinatorial Interaction Testing}

CIT stems from covering arrays, ultimately derived from the statistical field of design of experiments, and is principally concerned with designing tests that guarantee all interactions up to a certain level. In CIT, an interaction level is the number of system components for which possible interactions should be included in the test set. For example pairwise interaction testing, which is an interaction level of two, aims to design a test set with datapoints containing every possible interaction between the values of every two system components. A thorough review of CIT is provided by Nie and Leung \cite{nie2011survey}.  

CIT has been applied to several fields but has found a plethora of success in software testing. The application of CIT to software testing has proven capable of fault detection while minimizing the test set size requirements, as a majority of failures can be attributed to the interaction between few parameters \cite{22621}.

The extension of CIT to machine learning involves treating the feature space being used to train and test the model as the system parameters. Values are then the specific values each feature can take. A $t$-way interaction is the same as a $t$-way value combination. This is defined as a $t$-tuple of (feature, value) pairs. For example, a 3-way value combination for a car condition classification dataset could be a specific combination of values for mileage, age, and days since last inspection, e.g., `150,000 miles traveled', `20 years old', and `168 days since last inspection'.   

An extension of CIT is CC, which is the proportion of possible t-way interactions which appear in a set \cite{kuhn2013combinatorial}. `Covered' combinations are those interactions which do appear in a set, and `not covered' are those which do not appear in a set. As CC is concerned with the universe of all possible interactions, Lanus et al. extend CC to Set Difference Combinatorial Coverage (SDCC) \cite{lanus2021combinatorial}. SDCC is the proportion of interactions contained in one dataset but not in another, and is formally defined as follows.
\begin{definition} [$t$-way Set Difference Combinatorial Coverage]
    Let $D_L$ and $D_U$ be sets of data, and $D{_L}^t$ and $D{_U}^t$ be the corresponding $t$-way sets of data. The set difference $D{_U}^t \setminus D{_L}^t$ gives the value combinations that are in $D{_U}^t$ but that are not in $D{_L}^t$. The $t$-way set difference combinatorial coverage is
    $$SDCC^t(D_U, D_L) = \frac{|D{_U}^t \setminus D{_L}^t|}{|D{_U}^t|}.$$
\end{definition}  


Kuhn et al.\cite{kuhn2020combinatorial} show how combinatorial interactions can be used for explainable artificial intelligence. Combinatorial interactions has been used to better define the activity of hidden layers in deep learning \cite{ma2019deepct}, and CC has been used for the testing of deep learning models \cite{pei2017deepxplore,ma2018deepgauge}. CC has also been used as a holistic approach for training and testing of models \cite{cody2022systematic}. Lanus et al. utilize SDCC for failure analysis of machine learning, and find that a dataset with greater coverage leads to a better performing model \cite{lanus2021combinatorial}. Cody et al. expand their experiments and apply SDCC to MNIST \cite{cody2022systematic}.

\section{Methods}


The proposed query criterion relies heavily on SDCC, where the labeled dataset is considered $D_L$ and the unlabeled dataset $D_U$. The query strategy involves finding those datapoints in the unlabeled pool which contain interactions not included in the labeled set upto an interaction level of 6 as upto this level is where a majority of software failures are found \cite{22621}. Those datapoints which contain a greater number of missing interactions are to have a higher priority for labeling. Once the hierarchy of datapoints to label has been determined, selection according to this hierarchy is done in three ways; coverage density sampling, informative coverage density sampling, and uncertainty sampling weighted by coverage density. These data-centric methods should aid in sampling points which allow for data transferability to new models as illustrated in Figure 1.

Algorithm 1 presents a method to determine coverage density given unlabeled and labeled datasets. As a data point from the query set can contain several missing interactions, the sum of the number of missing interactions it contains could be considered as the density of coverage at that point. Lower level interactions are expected to be associated with a greater number of classes than higher level interactions, so they should hold a greater weight. The weighting scheme that is proposed utilizes the decreasing function $\frac{1}{t}$ for $t=1,...,6$ where each t is the t interaction level. The coverage density of some point is then the weighted sum of all interactions contained in that data point. This density is used in determining which datapoints to query in the proposed methods. Coverage density is formally defined as follows.

\begin{definition} [Coverage Density]
Let $D_L$ and $D_U$ be sets of data, and let $j \in D_L$ and $i \in D_U$. Also, let $j_t$ and $i_t$ be corresponding t-way set of data. Then coverage density of i at level t is 
$c_{i_t}$ = $\sum_{j \in D_L}{\frac{1}{t}\ \forall\ i_t\ \text{not in}\ j_t}$\\
Coverage Density of each $i \in D_U$ is
$\sum_{t \in T} c_{i_t} $, where T, the highest interaction level, is user specified.

\end{definition}

\begin{algorithm}
    \SetKwInOut{Input}{Input}
    \SetKwInOut{Output}{Output}
    \underline{function Coverage Density} $(L,U)$\;
    \Input{labeled Set $L$ and Unlabeled Set $U$}
    \Output{Coverage Density, $c$}
    \For([This loop finds all interactions at t in L and U]){$t$ $in$ $1$ $to$ $T$}{$\mu_t \gets \text{LIST of interactions in L at level t} $
    
    $\beta_t \gets \text{LIST of interactions in U at level t}$
    
    $indices \gets EMPTY LIST$
    
    \For([Update indices with missing interactions]){$i$ $in$ $1$ $to$ \text{length of} $\beta_t$}{
    \If{$i^{th}$ $element$ $of$ $\beta_t$ $not$ $in$ $\mu_t$}
      {
        APPEND associated index value from U to indices\;
      }
    }
    \For([Coverage Density is Updated]){$j$ $in$ $indices$}{
    $j^{th}$ $element$ $of$ $c$ $\gets$ $j^{th}$ $element$ $of$ $c + 1/t$
    }

    }
    \caption{Coverage Density Algorithm}
\end{algorithm}

All proposed sampling methods use the same definition of variables. $x_i$ is a binary variable valued 1 if data point $i$ will be sent to the oracle for labeling and 0 otherwise. $c_i$ is the coverage density of data point i as previously defined, and $b$ is the budget or number of points we are allowed to select. The three proposed methods to sample points are presented in Definitions 3-5.

\begin{definition} [Coverage Density Sampling]
Let $c_i$ and $b$ be given, the points selected are those with the highest coverage density:
            \begin{equation*}
                \begin{aligned}
                    \argmax_{i} \quad & \sum_{i}{ c_{i}x_i}\\
                    \textrm{s.t.} \quad & \sum_{i}x_{i} \leq b\\
                \end{aligned}
            \end{equation*}     
\end{definition}  

The second method weighs coverage density by similarity, which should protect against outliers. Cosine similarity is used as the measure of similarity between each query point and all other points, this is defined in Equation 1 where $\boldsymbol{x}$ represents all datapoints in both training and query sets. 

\begin{definition} [Informative Coverage Density Sampling]
Let $c_i$ and $b$ be given, the informativeness of a datapoint, $x_i$, is coverage density by similarity, where U is the cardinality of the unlabeled set: 
            \begin{equation*}
                \begin{aligned}
                    I(x_i) = c_i \frac{1}{U}sim(\boldsymbol{x},x_i)  
                \end{aligned}
            \end{equation*}
Where similarity is cosine similarity:
            \begin{equation}
                \begin{aligned}
                    sim(\boldsymbol{x},x_i) = \sum_{x \in \boldsymbol{x}}\frac{x\cdot x_i}{\left|x\right|\left|x_i\right|}
                \end{aligned}
            \end{equation}
The datapoints selected should maximize the sum of informativeness:      
            \begin{equation*}
                \begin{aligned}
                    \argmax_i \quad &  \sum_{i}{I(x_i)}\\
                    \textrm{s.t.} \quad & \sum_{i}x_{i} \leq b\\
                \end{aligned}
            \end{equation*}
\end{definition}

The final method involves weighting the common uncertainty sampling with entropy formulation by the coverage density of the data point as defined previously. 
\begin{definition} [Uncertainty Sampling Weighted by Coverage Density(USWCD)]
Let $c_i$ and $b$ be given, the informativeness of a data point, $x_i$, is defined as the following:
            \begin{equation*}
                \begin{aligned}
                    I(x_i) = H(x_i)c_{i}
                \end{aligned}
            \end{equation*}
Where $H(x_i)$ is the entropy of the model at prediction at point $x_i$:
            \begin{equation*}
                \begin{aligned}
                    H(x_i)= - \sum_{y \in Y}{p(y_i)log(p(y_i))}  
                \end{aligned}
            \end{equation*}
That is, the entropy over all the classes that a specific data point may belong too. 
The datapoints selected should maximize the sum of informativeness:
            \begin{equation*}
                \begin{aligned}
                    \argmax_i \quad &  \sum_{i}{I(x_i)}\\
                    \textrm{s.t.} \quad & \sum_{i}x_{i} \leq b\\
                \end{aligned}
            \end{equation*}
\end{definition}

These three methods are reliant on the data, with USWCD being the only method which takes some model input. Data-centric methods should allow for the data to be better transferred between models; this is illustrated in the experiments.  

\section{Experimental Design}
\label{sec:ed}
\subsection{Data}
All experiments are conducted on data sets from the UCI Machine Learning repository \cite{Dua:2019}, and the benchmark methods used are the ones previously defined. Table 1 displays general information about each data set. Batch size is the number of datapoints queried at each active learning iteration, for larger data sets a batch size of 100 is used while 25 points per sample is used for smaller data sets. All data sets, other than the Monk data set, are randomly split so that there are 10\% to test on. Of the remaining 90\% of data, 2.5\% is used as initial training data and 97.5\% is used as the query set. The Monk data set is pre-partitioned into training and testing sets, so the training set (which is the size listed in Table 1) is split into 97.5\% query points and 2.5\% initial training set.

\begin{table}[h!]
\captionsetup{justification=centering, labelsep=newline, font=footnotesize}
\centering
\caption{\sc Data Set Information}
\resizebox{9cm}{!}{%
\begin{tabular}{l c c c c} 
 \hline
 Data Set & datapoints & Features & Batch Size & Number of Batches  \\ [0.5ex] 
 \hline
 Tic-Tac-Toe & 957 & 9 & 100 & 8 \\ \hline
 Balance Scale & 624 & 4 & 25 & 21\\\hline
 Car Evaluation & 1727 & 6 & 100 & 15\\\hline
 Chess & 28066 & 6 & 100 & 246\\\hline
 Nursery & 12959 & 8 & 100 & 113\\\hline
 Monk & 414 & 6 & 25 & 13\\
 \hline
\end{tabular}}
\label{table:1}
\end{table}
\subsection{Performance Measures}
F1 is used as the measure of performance for each of the classifiers, as F1 will take into account class imbalance as well as model performance unlike model accuracy which only looks at model performance. Also, F1 balances precision and recall as opposed to other F-measures. F1 is calculated as 
\begin{equation*}
    \begin{aligned}
        F1 = 2* \frac{Precision*Recall}{Precision+Recall} \\
    \end{aligned}
\end{equation*}
Precision and Recall are defined as the following, where tp is true positive and fn is false negative:
\begin{equation*}
    \begin{aligned}
        Precision = \frac{tp}{tp+fp}  \\
        Recall = \frac{tp}{tp+fn}  \\
    \end{aligned}
\end{equation*}
Experiments on each data set are conducted three times, each time using the same random partition of data. The average F1 of the three runs is used to determine performance.

To quantify the performance over all iterations of sampling we take the area under the learning curve (AUC) for which the x-axis is number of datapoints queried and the y-axis is F1. The area is determined using the Trapezoidal rule as implemented in Numpy \cite{harris2020array}.

To determine the effect of the proposed methods on sampling bias, a sampling bias comparison method proposed by Krishan et al. \cite{krishnan2021mitigating} is used, this is presented in equation 2. $H_{D_{L}}$ is the entropy of the sampled set and $H_{Balanced}$ is the entropy of a set with an equal number of datapoints from each class.   

\begin{equation}
    \begin{aligned}
        \text{Sampling Bias} = 1 - \frac{H_{D_{L}}}{H_{Balanced}}
    \end{aligned}
\end{equation} 

$H_{D_{L}}$ is defined as $- \sum_{k=1}^{K}\frac{M_k}{M}log(\frac{M_k}{M})$ where $M_k$ is the number of datapoints belonging to class k and $M$ is the total number of datapoints in our sample.  The average of sampling bias of the three runs is used for comparison. 

\subsection{Experiment 1}
The initial experiment involves sampling with respect to a certain model and using the same model to compare methods. This is the active learning scenario most frequently discussed and presented in the `initial labeling effort' section of Figure 1.  All data sets are sampled with respect to, trained, and tested utilizing a Random Forest Classifier with max depth constrained to five.

\subsection{Experiment 2}
To determine the effectiveness of the methods in sampling points that are beneficial to a model outside the learning loop as depicted in the `Reuse of Labeled Data' portion of Figure 1, another experiment is employed. The points sampled with respect to a Random Forest Classifier with max depth constrained to 5 are used to train a Decision Tree classifier and Support Vector Machine (SVM); for the SVM a Support Vector Classifier (SVC) implementation is used. Both the Decision Tree and SVC do not have any hyperparameter tuning and are utilized as is from Scikit-learn \cite{scikit-learn}. That is, the Decision Tree uses Gini impurity to determine quality of split and is not contrained to a max depth, the SVC uses a radial basis function kernel and has the squared L2 norm as a regularization term. After each iteration of sampling, all models are tested and the average of the three runs is used for the results. These experiments should provide an understanding as to how methods compare when sampling and training with respect to a particular model, and further will depict the effectiveness of these sampled points to transfer to different models. Both scenarios are crucial to machine learning and active learning deployment. 

\section{Results}
\begin{figure*}[h]
     \centering
     \begin{subfigure}[b]{0.45\textwidth}
         \centering
         \includegraphics[width=\textwidth,height=5cm,keepaspectratio]{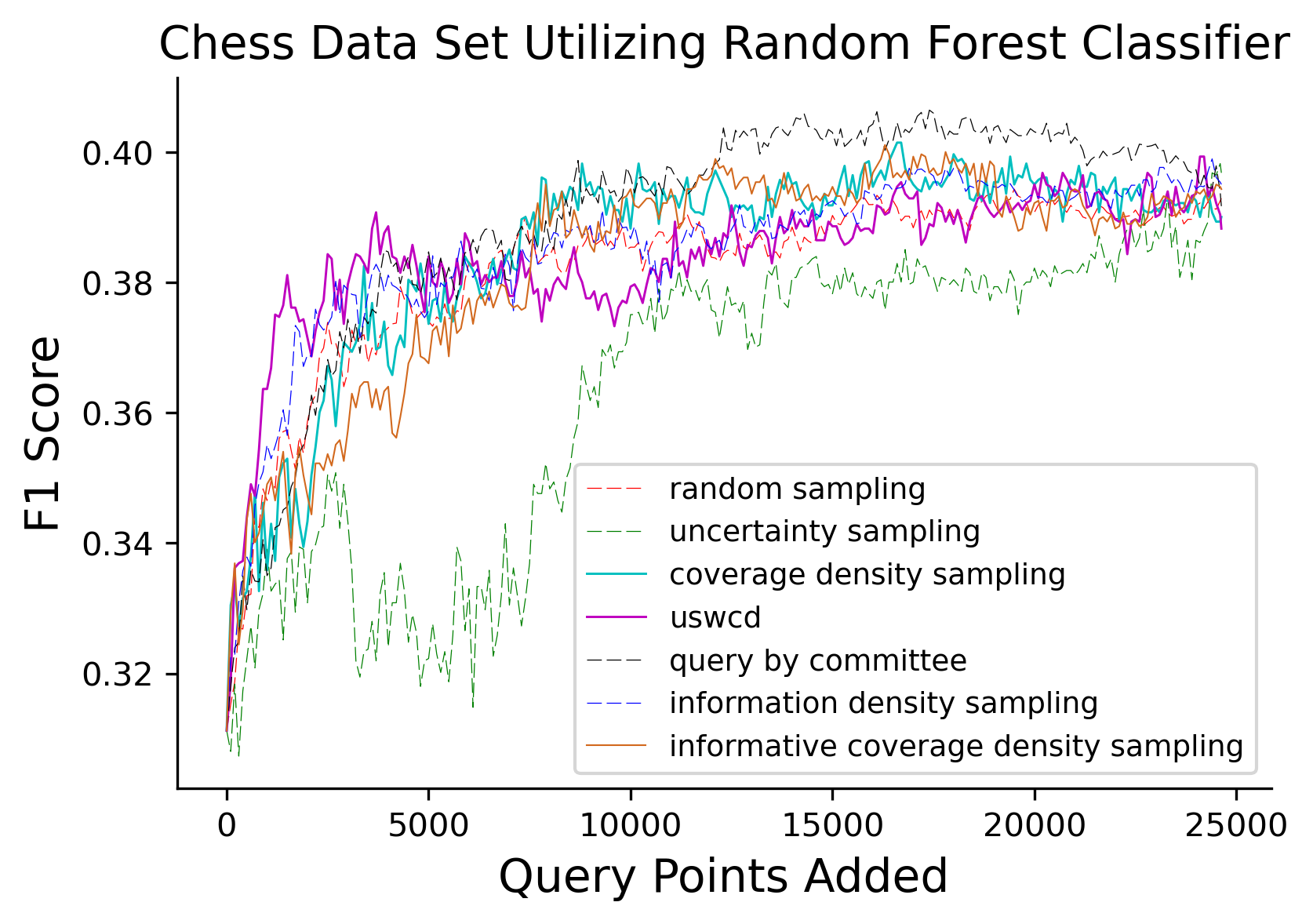}
         \caption{Chess with Random Forest}
         \label{fig:chess_forest}
     \end{subfigure}
     \begin{subfigure}[b]{0.45\textwidth}
         \centering
         \includegraphics[width=\textwidth,height=5cm,keepaspectratio]{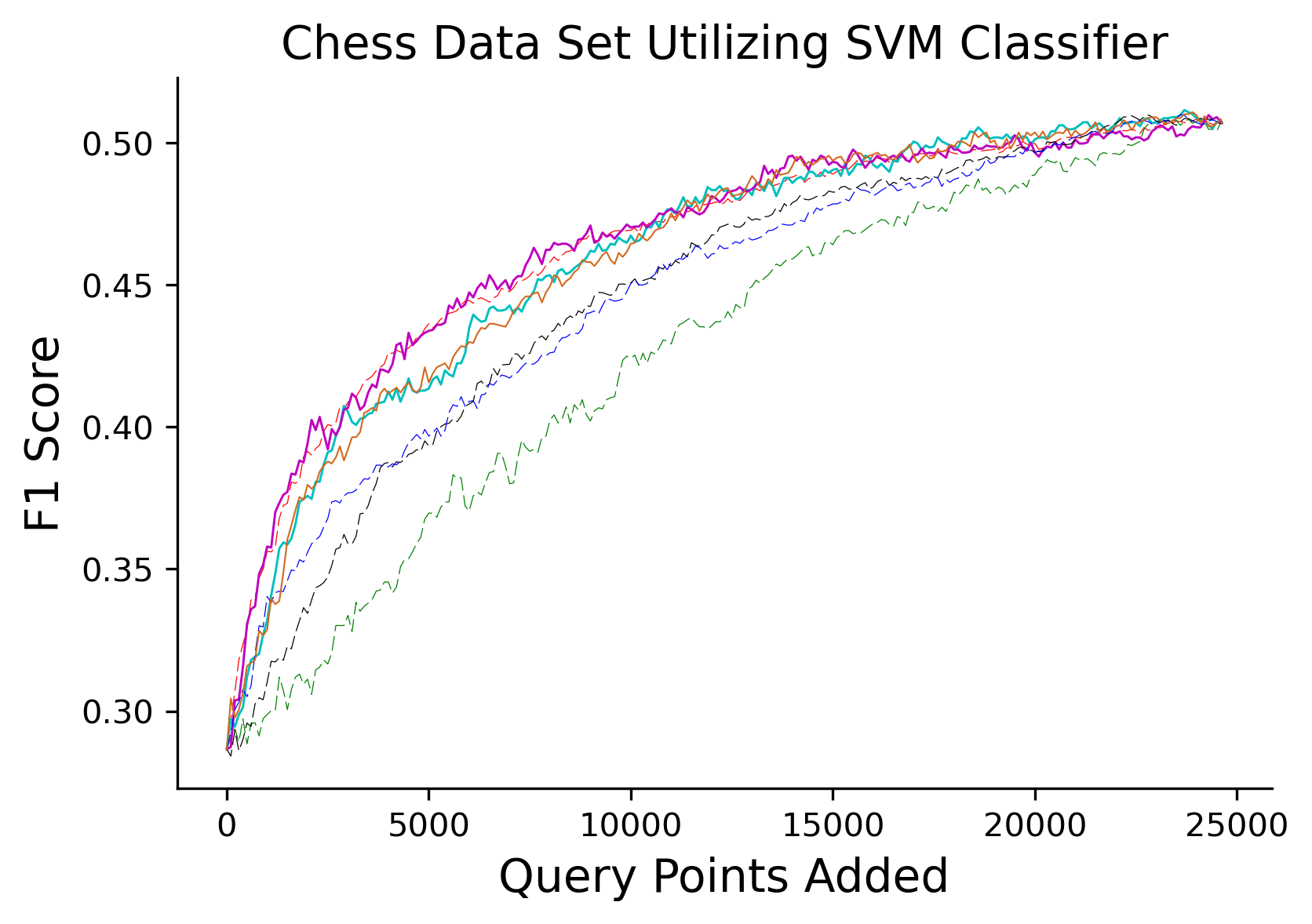}
         \caption{Chess with SVM}
         \label{fig:chess_svm}
     \end{subfigure}
        \caption{Chess Dataset Results}
        \label{fig:chess_results}
\end{figure*}
\begin{figure*}[h]
     \centering
     \begin{subfigure}[b]{0.45\textwidth}
         \centering
         \includegraphics[width=\textwidth,height=5cm,keepaspectratio]{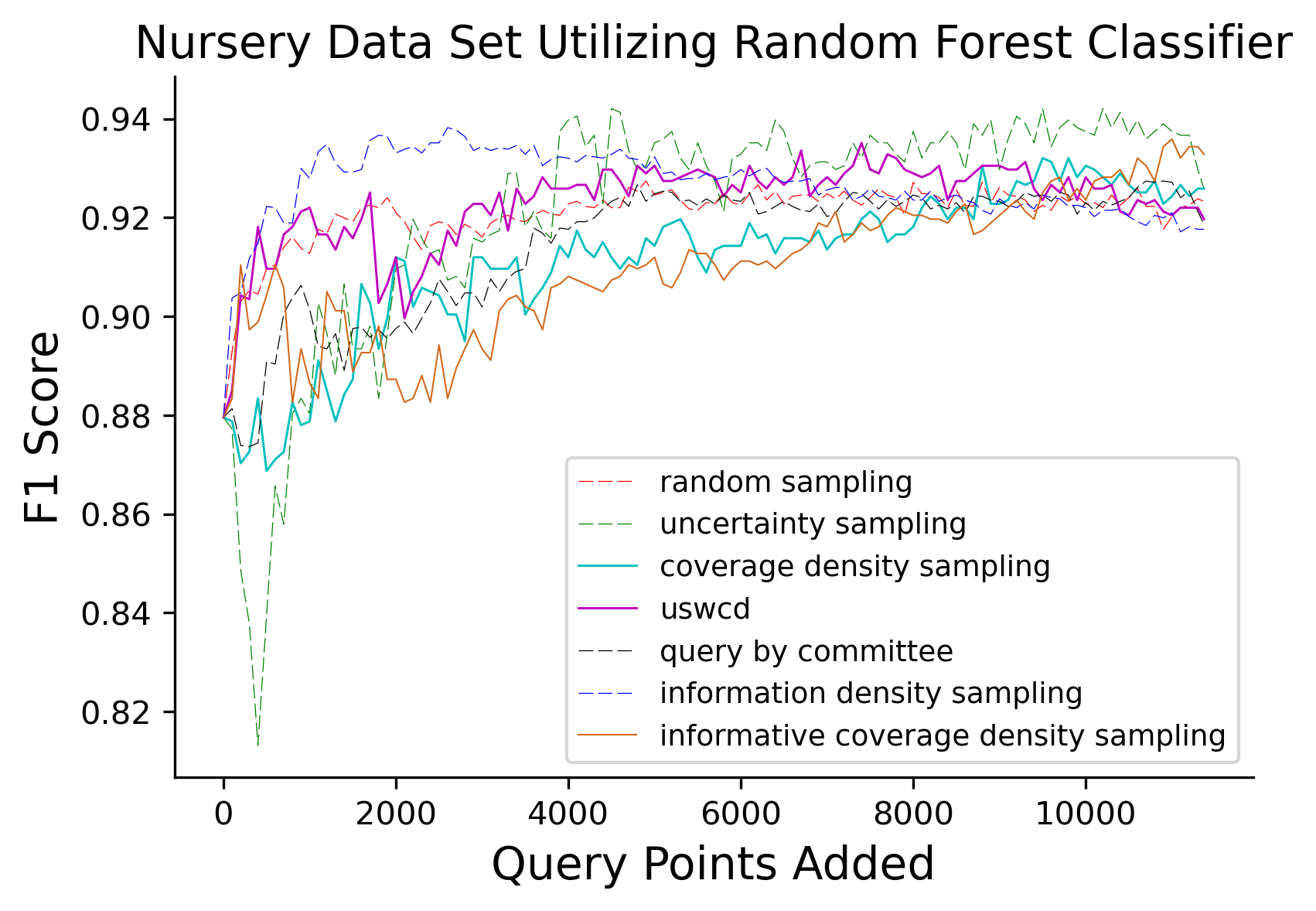}
         \caption{Nursery with Random Forest}
         \label{fig:nursery_forest}
     \end{subfigure}
     \begin{subfigure}[b]{0.45\textwidth}
         \centering
         \includegraphics[width=\textwidth,height=5cm,keepaspectratio]{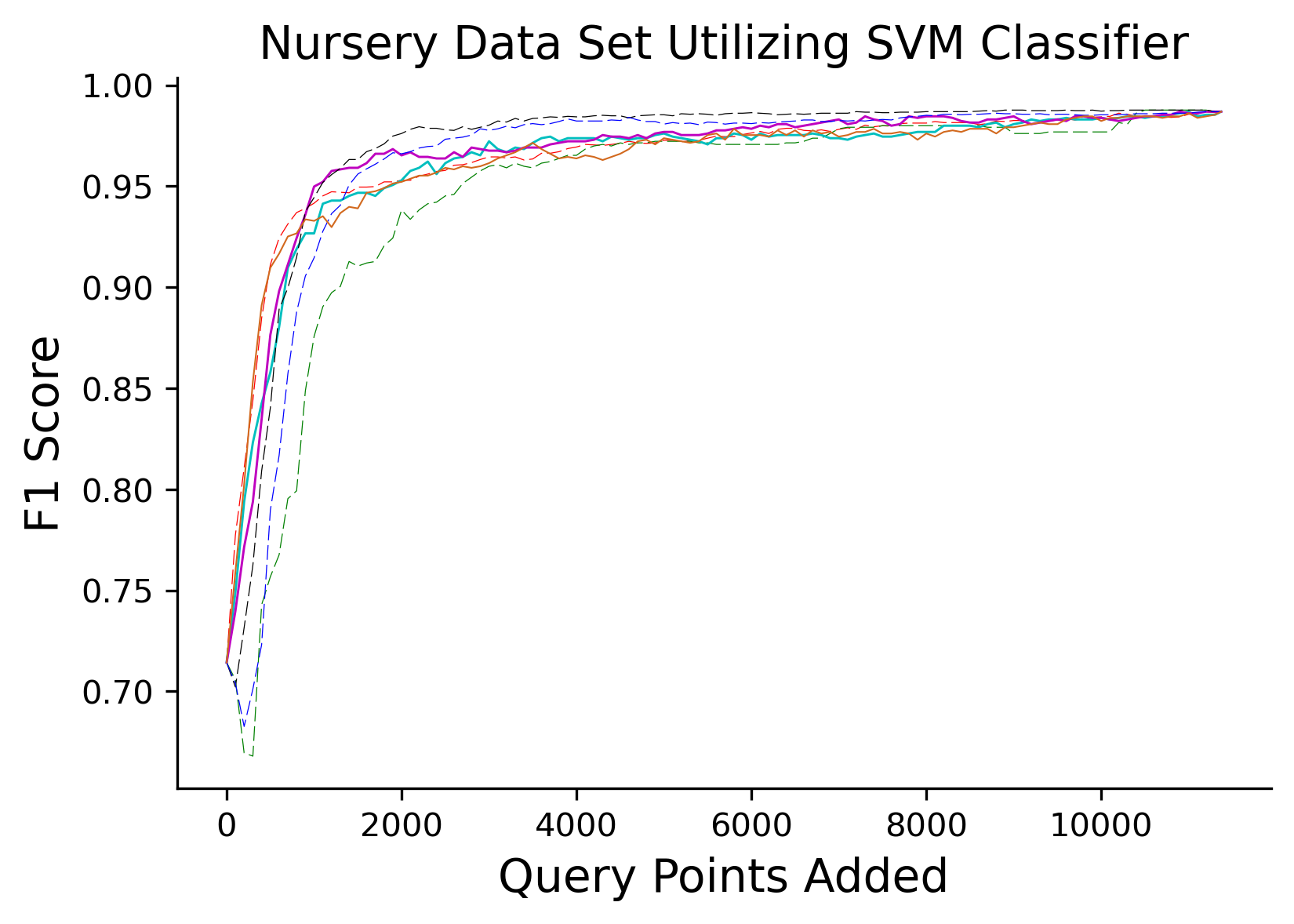}
         \caption{Nursery with SVM}
         \label{fig:nursery_svm}
     \end{subfigure}
        \caption{Nursery Dataset Results}
        \label{fig:nursery_results}
\end{figure*}

The learning curves for the chess and nursery datasets using Random Forest and SVM classifiers are in figures 2 and 3; these plots are F1 score versus number of query points added. Dashed lines signify the benchmark methods while solid lines signify the methods incorporating coverage. AUC of F1 vs query points added to training set for each of the sampling methods are presented in Table 2. Query by Committee has the best performance a majority of the trials, seven, but USWCD is a near second with six best performances. To break down instances in which each method outperforms the others we look at the performance on the original model as well as on the models that the data is transferred to.

We first focus on `Experiment 1' as described in Section \ref{sec:ed}. The proposed methods should be competitive with the benchmark methods when sampling and testing with respect to a specific model, in our case the Random Forest model restricted to a depth of five. To examine this performance we look at the percent difference in AUC of each method from the best performing method on each dataset, this is presented in Table 3. USWCD is the best performer once, it has 0.00\% difference in AUC from the best method, itself. Uncertainty sampling and information density also perform best only once, but Query by Committee performs best three times. Though USWCD does not always perform the best, it achieves performance nearest the best performer in 60\% of instances where it is not the best performer. So, USWCD does achieve a competitive performance for the model in the active learning loop, next we study models outside the learning loop as described in `Experiment 2'.   

In machine learning deployment, the model in use would likely not change unless the new model provides some benefit such as computational efficiency or performance. For this study we pay special attention to performance, and look at instances in which the use of a Decision Tree or SVM increases final model performance by 5\% or more. As Lowell et al. \cite{lowell2019practical} point out, active learning methods often do not outperform random sampling when the sampled data is transferred to a new model. Therefore, we treat random sampling as a baseline for data transfer comparison. Table 4 shows the percent difference in area under F1 versus queried points curve between each method and random sampling when model performance increases by 5\% or more. In a majority of the presented scenarios USWCD outperforms the other methods. This is also the only method which does not perform worse than random sampling in any instance of model improvement. So, the proposed method is effective in sampling data which is transferable to new and more effective models.

\begin{table*}[t]
\captionsetup{justification=centering, labelsep=newline, font=footnotesize}
\centering
  \caption{\sc AUC of F1 vs. Query Points Added For Each Method, Model, and Dataset}
  \begin{tabular}{llllllll}
     Active Learning Method & Model & Monk & Balance Scale & Car Evaluation & Tic-Tac-Toe & Nursery & Chess \\ \hline
    Random Sampling & \multirow{3}{7em}{Random Forest\\ Decision Tree \\ SVM} & \multirow{3}{4em}{221.60\\210.46 \\ 219.71} & \multirow{3}{4em}{462.63\\426.47 \\ 488.30}& \multirow{3}{4em}{1303.77\\1509.68 \\ 1321.41} & \multirow{3}{4em}{696.84\\764.03 \\ 690.35}&\multirow{3}{4em}{10499.40\\11201.24 \\ 10997.11}&\multirow{3}{4em}{9437.38\\17051.97 \\ 11464.67}
    \\
    \\
    \\
    \\ \hline
    Uncertainty Sampling & \multirow{3}{7em}{Random Forest\\ Decision Tree \\ SVM}& \multirow{3}{4em}{\textbf{228.47}\\213.20 \\ 221.67} & \multirow{3}{4em}{459.87\\\textbf{433.73} \\ 477.82} & \multirow{3}{4em}{1332.55\\1455.13 \\ 1305.52}& \multirow{3}{4em}{714.21\\758.59 \\ 710.00} & \multirow{3}{4em}{10507.02\\11126.62 \\ 10800.42}&\multirow{3}{4em}{8972.98\\15848.8 \\ 10537.07} \\
    \\
    \\
    \\ \hline
    Query by Committee & \multirow{3}{7em}{Random Forest\\ Decision Tree \\ SVM}& \multirow{3}{4em}{224.90\\\textbf{217.30}\\222.03} & \multirow{3}{4em}{\textbf{477.55}\\415.52 \\ 489.51}& \multirow{3}{4em}{1346.31\\1491.37 \\ 1395.44}& \multirow{3}{4em}{\textbf{731.57}\\761.22 \\ \textbf{717.71}}& \multirow{3}{4em}{10430.06\\\textbf{11301.59} \\ \textbf{11056.61}}&\multirow{3}{4em}{\textbf{9632.39}\\17196.40 \\ 11031.15}\\
    \\
    \\
    \\ \hline   
    Information Density Sampling & \multirow{3}{7em}{Random Forest\\ Decision Tree \\ SVM}& \multirow{3}{4em}{220.91\\213.28 \\ 218.88}& \multirow{3}{4em}{466.80\\419.28 \\ 488.57}& \multirow{3}{4em}{1374.51\\1517.34 \\ 1403.19}& \multirow{3}{4em}{712.80\\760.35 \\ 703.33}& \multirow{3}{4em}{\textbf{10563.24}\\11299.12 \\ 10971.20}& \multirow{3}{4em}{9511.56\\\textbf{17593.14} \\ 11048.85}  \\
    \\
    \\
    \\ \hline   
    Coverage Density Sampling & \multirow{3}{7em}{Random Forest\\ Decision Tree \\ SVM}& \multirow{3}{4em}{219.68\\211.20\\ \textbf{222.81}}& \multirow{3}{4em}{453.42\\408.87 \\ 484.27} & \multirow{3}{4em}{1297.09\\1507.07 \\ 1304.94}& \multirow{3}{4em}{705.78\\775.08 \\ 681.57}& \multirow{3}{4em}{10390.65\\11090.34 \\ 10969.15}& \multirow{3}{4em}{9512.64\\16704.74 \\ 11381.05}  \\
    \\
    \\
    \\ \hline
    Informative Coverage Density Sampling & \multirow{3}{7em}{Random Forest\\ Decision Tree \\ SVM}& \multirow{3}{4em}{222.02\\209.71\\ 221.73}& \multirow{3}{4em}{445.56\\409.81 \\ 479.83}& \multirow{3}{4em}{1314.82\\1505.03 \\ 1317.73}& \multirow{3}{4em}{693.15\\759.47 \\ 686.84}& \multirow{3}{4em}{10376.79\\11065.89 \\ 10970.84}& \multirow{3}{4em}{9465.12\\16720.74 \\ 11369.16}  \\
    \\
    \\
    \\ \hline   
    USWCD & \multirow{3}{7em}{Random Forest\\ Decision Tree \\ SVM}& \multirow{3}{4em}{220.05\\209.74\\218.84}& \multirow{3}{4em}{477.21\\425.33 \\ \textbf{493.54}}& \multirow{3}{4em}{\textbf{1388.95}\\\textbf{1531.87} \\ \textbf{1404.94}}& \multirow{3}{4em}{716.84\\\textbf{778.59} \\ 711.05} & \multirow{3}{4em}{10522.23\\11245.97\\ 11006.75}& \multirow{3}{4em}{9485.49\\17300.60 \\ \textbf{11482.56}}  \\ 
    \\
    \\
    \\
    \hline
  \end{tabular}
\end{table*}

  

\begin{table*}[t]
\captionsetup{justification=centering, labelsep=newline, font=footnotesize}
  \centering
  \caption{\sc Percent Difference in AUC Between Best Method And Each Method on Original Model by Dataset (\%)}
  \begin{tabular}{lcccccc}
  Active Learning Method & Monk & Balance Scale & Car Evaluation & Tic-Tac-Toe & Nursery & Chess \\ \hline
  Random Sampling & -3.01 &-3.13 &-6.13 & -4.75 &-0.61 &-2.03 \\ \hline
  Uncertainty Sampling & \textbf{0.00} &-3.70 &-4.06 &-2.37 &-0.53 &-6.85 \\ \hline
  Query by Committee & -1.56 &\textbf{0.00} &-3.07 &\textbf{0.00}&-1.26 &\textbf{0.00} \\ \hline
  Information Density Sampling & -3.31&-2.25&-1.04&-2.57&\textbf{0.00}&-1.25 \\ \hline
  Coverage Density Sampling & -3.85&-5.05&-6.61&-3.53&-1.63&-1.24\\ \hline 
  Informative Coverage Density Sampling & -2.82&-6.70&-5.34&-5.25&-1.77&-1.74 \\ \hline
  USWCD & -3.69&-0.07&\textbf{0.00}&-2.01&-0.39&-1.53\\ \hline
  \end{tabular}
  
\end{table*}

\begin{table*}[t]
\captionsetup{justification=centering, labelsep=newline, font=footnotesize}
\centering
\caption{\sc Percent Difference in AUC from Random Sampling When Model Changes and Performance Increases (\%)}
  \begin{tabular}{lllllll}
     Active Learning Method & Model  & Balance Scale & Car Evaluation & Tic-Tac-Toe & Nursery & Chess \\ \hline
    Uncertainty Sampling & \multirow{2}{7em}{Decision Tree \\ SVM}& \multirow{2}{5em}{ \\ -2.15} & \multirow{2}{5em}{-3.61 \\ -1.20} & \multirow{2}{5em}{\hspace{.14cm}-.71\\}& \multirow{2}{5em}{\hspace{.14cm}-.67\\-1.79} & \multirow{2}{5em}{-7.06\\-8.09}
    \\
    \\
    \hline
    Query by Committee & \multirow{2}{7em}{Decision Tree \\ SVM}& \multirow{2}{5em}{\\\hspace{.22cm}.25} & \multirow{2}{5em}{-1.21\\\hspace{.09cm}5.60}& \multirow{2}{5em}{\hspace{.14cm}-.37 \\}& \multirow{2}{5em}{\hspace{.23cm}\textbf{.9}\\ \hspace{.22cm}\textbf{.54}}& \multirow{2}{5em}{\hspace{.22cm}.85\\-3.78}
    \\
    \\
    \hline   
    Information Density Sampling & \multirow{2}{7em}{Decision Tree \\ SVM}& \multirow{2}{5em}{\\\hspace{.22cm}.06}& \multirow{2}{5em}{\hspace{.22cm}.51\\\hspace{.09cm}6.19}& \multirow{2}{5em}{\hspace{.14cm}-.48\\}& \multirow{2}{5em}{\hspace{.22cm}.87\\\hspace{.14cm}-.24}& \multirow{2}{5em}{\hspace{.09cm}\textbf{3.17}\\-3.63} 
    \\
    \\
    \hline   
    Coverage Density Sampling & \multirow{2}{7em}{Decision Tree \\ SVM}& \multirow{2}{5em}{\\\hspace{.14cm}-.83}& \multirow{2}{5em}{\hspace{.14cm}-.17\\-1.25} & \multirow{2}{5em}{\hspace{.09cm}1.45\\}& \multirow{2}{5em}{\hspace{.14cm}-.99\\\hspace{.14cm}-.25}& \multirow{2}{5em}{-2.04\\\hspace{.14cm}-.73}  
    \\
    \\
    \hline
    Informative Coverage Density Sampling & \multirow{2}{7em}{Decision Tree \\ SVM}& \multirow{2}{5em}{\\-1.74}& \multirow{2}{5em}{\hspace{.14cm}-.31\\\hspace{.14cm}-.28}& \multirow{2}{5em}{\hspace{.14cm}-.6\\}& \multirow{2}{5em}{-1.21\\\hspace{.14cm}-.24}& \multirow{2}{5em}{-1.94\\\hspace{.14cm}-.83}  
    \\
    \\
     \hline   
    USWCD & \multirow{2}{7em}{Decision Tree \\ SVM}& \multirow{2}{5em}{\\\hspace{.09cm}\textbf{1.07}}& \multirow{2}{5em}{\hspace{.09cm}\textbf{1.47}\\\hspace{.09cm}\textbf{6.32}}& \multirow{2}{5em}{\hspace{.09cm}\textbf{1.91} \\}& \multirow{2}{5em}{\hspace{.22cm}.40\\\hspace{.22cm}.09} & \multirow{2}{5em}{\hspace{.09cm}1.46\\\hspace{.22cm}\textbf{.16}} \\ 
    \\
    \hline
  \end{tabular}
\end{table*}


The proposed USWCD does achieve comparable performance to the benchmark methods on the original model, and outperforms the benchmark methods on new models which perform better. However, comparing methods across all datasets for each model would help to paint a better picture of overall performance. To compare results of active learning methods across datasets, the area under the F1 curve is normalized using the following equation: $ x^* = \frac{x-x_{min}}{x_{max}-x_{min}}$.

Then, the median performance of each method across all datasets is recorded, these results are presented in Table 5, where the top performer is bolded. It is clear that the proposed USWCD method outperforms the other methods when data is transfered to new models. On the original model, query by committee performs best but USWCD achieves the most similar performance. So, UWSCD achieves a comparable performance to existing methods in the single-model setting, and outperforms existing methods in the multi-model setting.    

\begin{table*}[t]
\captionsetup{justification=centering, labelsep=newline, font=footnotesize}
  \centering
  \caption{\sc Median Normalized Area Under Curve of F1 vs. Query Points Added}
  \begin{tabular}{lccc}
  Active Learning Method & Random Forest & Decision Tree & SVM \\ \hline
  Random Sampling & .376 & .632 & .455\\\hline
  Uncertainty Sampling & .498 & .129 & .003\\\hline
  Query by Committee & \textbf{.797} & .622 & .854 \\\hline
  Information Density Sampling & .740 & .641 & .634\\\hline
  Coverage Density Sampling & .160 & .344 & .534 \\\hline 
  Informative Coverage Density Sampling & .097 & .041 & .406 \\\hline
  USWCD & .779 & \textbf{.798} & \textbf{.908} \\\hline
  \end{tabular}
\end{table*}




To study the sampling bias across all datasets, a violin plot is used as implemented in seaborn\cite{Waskom2021}. The violin plot is created using kernel density estimation for fitting a probability distribution. The white dot in each distribution marks the median. The plot is shown in Figure 4, and the median sampling bias of USWCD is lower than that of all other methods. However, the distributions do look very similar, so testing the methods in datasets with a greater class diversity may help to paint a better picture of sampling bias.

\begin{figure}[t]
						\centering
						\includegraphics[scale=.4]{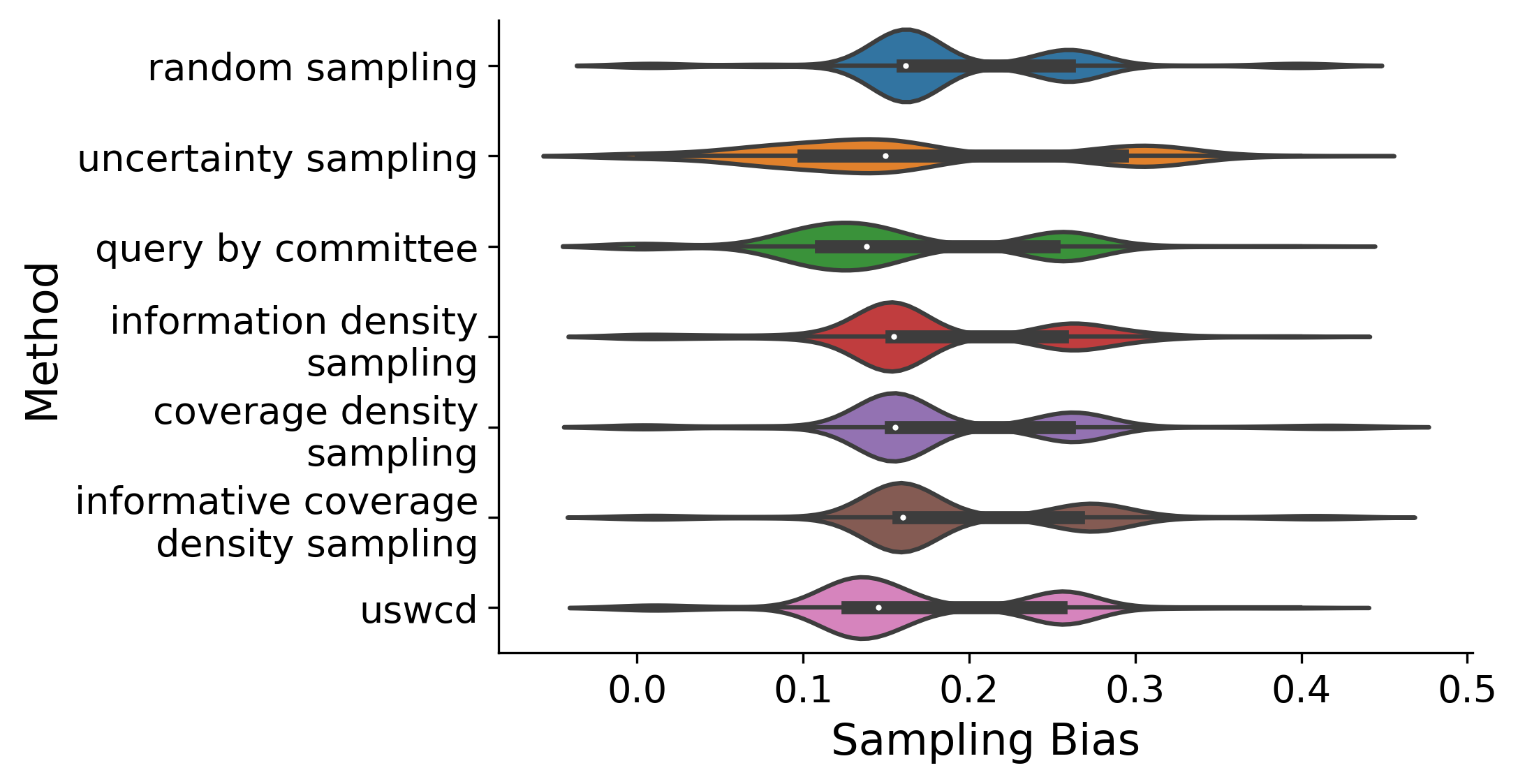}
						\caption{Violin Plot for Sampling Bias} 
						\label{fig:violin plot}
\end{figure}


\section{Discussion and Future Work}
As shown in the experiment results, our proposed USWCD achieves a comparable performance on the model in the active learning loop, and outperforms existing methods on models outside the loop. The versatility of the sampled data is crucial
for deployable active learning. 
We also study the effect on sampling bias as a data-centric approach may ease these issues usually associated with active learning. It is found that there are not large improvements in sampling bias compared to all benchmark methods, but the proposed USWCD does achieve a lower median sampling bias than other methods. Studying the methods on datasets with more classes may help to paint a better picture of the sampling bias.

Other future extensions of this work include the extension of  coverage based active learning to continuous data, where a specific discretization method would be applied to determine which datapoints contain missing interactions. The extension of coverage density sampling to neural networks, deep learning, and other computationally expensive models could also be beneficial. Re-training these models at each iteration of active learning, as is required for uncertainty sampling, query by committee, and information density sampling, might not be ideal. Coverage density could offer a data-centric approach to sample points without the need to re-train the model at each iteration.

\section{Conclusion}
In this paper we introduced three active learning methods utilizing SDCC with the goal of creating  methods that sample data which is useful not only for the current model but future models as well. We test the methods on six data sets from the UCI machine learning repository and use four benchmark methods: random sampling, uncertainty sampling, query by committee, and information density sampling. We initially sample data using a Random Forest classifier, then use the sampled data to train a Decision Tree and SVM to test our hypothesis that a data-centric method should sample datapoints which contribute more to learning for any arbitrary model than a heavily model focused method. The results are summarized in Table 5. We find that the proposed USWCD method performs similarly if not better than other methods on the original model. When the data is transferred to new models, USWCD outperforms all other methods. 



\bibliographystyle{IEEEtran}
\bibliography{ref}

\begin{thebibliography}{10}
\providecommand{\url}[1]{#1}
\csname url@samestyle\endcsname
\providecommand{\newblock}{\relax}
\providecommand{\bibinfo}[2]{#2}
\providecommand{\BIBentrySTDinterwordspacing}{\spaceskip=0pt\relax}
\providecommand{\BIBentryALTinterwordstretchfactor}{4}
\providecommand{\BIBentryALTinterwordspacing}{\spaceskip=\fontdimen2\font plus
\BIBentryALTinterwordstretchfactor\fontdimen3\font minus
  \fontdimen4\font\relax}
\providecommand{\BIBforeignlanguage}[2]{{%
\expandafter\ifx\csname l@#1\endcsname\relax
\typeout{** WARNING: IEEEtran.bst: No hyphenation pattern has been}%
\typeout{** loaded for the language `#1'. Using the pattern for}%
\typeout{** the default language instead.}%
\else
\language=\csname l@#1\endcsname
\fi
#2}}
\providecommand{\BIBdecl}{\relax}
\BIBdecl

\bibitem{polyzotis2018data}
N.~Polyzotis, S.~Roy, S.~E. Whang, and M.~Zinkevich, ``Data lifecycle
  challenges in production machine learning: a survey,'' \emph{ACM SIGMOD
  Record}, vol.~47, no.~2, pp. 17--28, 2018.

\bibitem{settles1995active}
B.~Settles, ``Active learning literature survey,'' \emph{Science}, vol.~10,
  no.~3, pp. 237--304, 1995.

\bibitem{yang2017suggestive}
L.~Yang, Y.~Zhang, J.~Chen, S.~Zhang, and D.~Z. Chen, ``Suggestive annotation:
  A deep active learning framework for biomedical image segmentation,'' in
  \emph{International conference on medical image computing and
  computer-assisted intervention}.\hskip 1em plus 0.5em minus 0.4em\relax
  Springer, 2017, pp. 399--407.

\bibitem{settles2008analysis}
B.~Settles and M.~Craven, ``An analysis of active learning strategies for
  sequence labeling tasks,'' in \emph{proceedings of the 2008 conference on
  empirical methods in natural language processing}, 2008, pp. 1070--1079.

\bibitem{hoi2006batch}
S.~C. Hoi, R.~Jin, J.~Zhu, and M.~R. Lyu, ``Batch mode active learning and its
  application to medical image classification,'' in \emph{Proceedings of the
  23rd international conference on Machine learning}, 2006, pp. 417--424.

\bibitem{zhao2013cost}
P.~Zhao and S.~C. Hoi, ``Cost-sensitive online active learning with application
  to malicious url detection,'' in \emph{Proceedings of the 19th ACM SIGKDD
  international conference on Knowledge discovery and data mining}, 2013, pp.
  919--927.

\bibitem{nissim2015boosting}
N.~Nissim, A.~Cohen, and Y.~Elovici, ``Boosting the detection of malicious
  documents using designated active learning methods,'' in \emph{2015 IEEE 14th
  International Conference on Machine Learning and Applications (ICMLA)}.\hskip
  1em plus 0.5em minus 0.4em\relax IEEE, 2015, pp. 760--765.

\bibitem{dasari2021active}
S.~K. Dasari, A.~Cheddad, L.~Lundberg, and J.~Palmquist, ``Active learning to
  support in-situ process monitoring in additive manufacturing,'' in \emph{2021
  20th IEEE International Conference on Machine Learning and Applications
  (ICMLA)}.\hskip 1em plus 0.5em minus 0.4em\relax IEEE, 2021, pp. 1168--1173.

\bibitem{lowell2019practical}
D.~Lowell, Z.~C. Lipton, and B.~C. Wallace, ``Practical obstacles to deploying
  active learning,'' in \emph{Proceedings of the 2019 Conference on Empirical
  Methods in Natural Language Processing and the 9th International Joint
  Conference on Natural Language Processing (EMNLP-IJCNLP)}, 2019, pp. 21--30.

\bibitem{pmlr-v16-tomanek11a}
\BIBentryALTinterwordspacing
K.~Tomanek and K.~Morik, ``Inspecting sample reusability for active learning,''
  in \emph{Active Learning and Experimental Design workshop In conjunction with
  AISTATS 2010}, ser. Proceedings of Machine Learning Research, I.~Guyon,
  G.~Cawley, G.~Dror, V.~Lemaire, and A.~Statnikov, Eds., vol.~16.\hskip 1em
  plus 0.5em minus 0.4em\relax Sardinia, Italy: PMLR, 16 May 2011, pp.
  169--181. [Online]. Available:
  \url{https://proceedings.mlr.press/v16/tomanek11a.html}
\BIBentrySTDinterwordspacing

\bibitem{paleyes2020challenges}
A.~Paleyes, R.-G. Urma, and N.~D. Lawrence, ``Challenges in deploying machine
  learning: a survey of case studies,'' \emph{arXiv preprint arXiv:2011.09926},
  2020.

\bibitem{dasgupta2008hierarchical}
S.~Dasgupta and D.~Hsu, ``Hierarchical sampling for active learning,'' in
  \emph{Proceedings of the 25th International Conference on Machine Learning},
  2008, pp. 208--215.

\bibitem{krishnan2021mitigating}
R.~Krishnan, A.~Sinha, N.~Ahuja, M.~Subedar, O.~Tickoo, and R.~Iyer,
  ``Mitigating sampling bias and improving robustness in active learning,''
  \emph{arXiv preprint arXiv:2109.06321}, 2021.

\bibitem{sener2018active}
O.~Sener and S.~Savarese, ``Active learning for convolutional neural networks:
  A core-set approach,'' in \emph{International Conference on Learning
  Representations}, 2018.

\bibitem{agarwal2020contextual}
S.~Agarwal, H.~Arora, S.~Anand, and C.~Arora, ``Contextual diversity for active
  learning,'' in \emph{European Conference on Computer Vision}.\hskip 1em plus
  0.5em minus 0.4em\relax Springer, 2020, pp. 137--153.

\bibitem{Liu_2021_ICCV}
Z.~Liu, H.~Ding, H.~Zhong, W.~Li, J.~Dai, and C.~He, ``Influence selection for
  active learning,'' in \emph{Proceedings of the IEEE/CVF International
  Conference on Computer Vision (ICCV)}, October 2021, pp. 9274--9283.

\bibitem{Elhamifar_2013_ICCV}
E.~Elhamifar, G.~Sapiro, A.~Yang, and S.~S. Sasrty, ``A convex optimization
  framework for active learning,'' in \emph{Proceedings of the IEEE
  International Conference on Computer Vision (ICCV)}, December 2013.

\bibitem{pei2017deepxplore}
K.~Pei, Y.~Cao, J.~Yang, and S.~Jana, ``Deepxplore: Automated whitebox testing
  of deep learning systems,'' in \emph{proceedings of the 26th Symposium on
  Operating Systems Principles}, 2017, pp. 1--18.

\bibitem{ma2018deepgauge}
L.~Ma, F.~Juefei-Xu, F.~Zhang, J.~Sun, M.~Xue, B.~Li, C.~Chen, T.~Su, L.~Li,
  Y.~Liu \emph{et~al.}, ``Deepgauge: Multi-granularity testing criteria for
  deep learning systems,'' in \emph{Proceedings of the 33rd ACM/IEEE
  International Conference on Automated Software Engineering}, 2018, pp.
  120--131.

\bibitem{ma2019deepct}
L.~Ma, F.~Juefei-Xu, M.~Xue, B.~Li, L.~Li, Y.~Liu, and J.~Zhao, ``Deepct:
  Tomographic combinatorial testing for deep learning systems,'' in \emph{2019
  IEEE 26th International Conference on Software Analysis, Evolution and
  Reengineering (SANER)}.\hskip 1em plus 0.5em minus 0.4em\relax IEEE, 2019,
  pp. 614--618.

\bibitem{kuhn2020combinatorial}
D.~R. Kuhn, R.~N. Kacker, Y.~Lei, and D.~E. Simos, ``Combinatorial methods for
  explainable {AI},'' in \emph{2020 IEEE International Conference on Software
  Testing, Verification and Validation Workshops (ICSTW)}.\hskip 1em plus 0.5em
  minus 0.4em\relax IEEE, 2020, pp. 167--170.

\bibitem{lanus2021combinatorial}
E.~Lanus, L.~J. Freeman, D.~R. Kuhn, and R.~N. Kacker, ``Combinatorial testing
  metrics for machine learning,'' in \emph{2021 IEEE International Conference
  on Software Testing, Verification and Validation Workshops (ICSTW)}.\hskip
  1em plus 0.5em minus 0.4em\relax IEEE, 2021, pp. 81--84.

\bibitem{cody2022systematic}
T.~Cody, E.~Lanus, D.~D. Doyle, and L.~Freeman, ``Systematic training and
  testing for machine learning using combinatorial interaction testing,'' in
  \emph{2022 IEEE International Conference on Software Testing, Verification
  and Validation Workshops (ICSTW)}.\hskip 1em plus 0.5em minus 0.4em\relax
  IEEE, 2022, pp. 102--109.

\bibitem{lewis1994heterogeneous}
D.~D. Lewis and J.~Catlett, ``Heterogeneous uncertainty sampling for supervised
  learning,'' in \emph{Machine Learning Proceedings}.\hskip 1em plus 0.5em
  minus 0.4em\relax Elsevier, 1994, pp. 148--156.

\bibitem{seung1992query}
H.~S. Seung, M.~Opper, and H.~Sompolinsky, ``Query by committee,'' in
  \emph{Proceedings of the Fifth Annual Workshop on Computational Learning
  Theory}, 1992, pp. 287--294.

\bibitem{baldridge2004active}
J.~Baldridge and M.~Osborne, ``Active learning and the total cost of
  annotation,'' in \emph{Proceedings of the 2004 Conference on Empirical
  Methods in Natural Language Processing}, 2004, pp. 9--16.

\bibitem{pardakhti2021practical}
M.~Pardakhti, N.~Mandal, A.~W. Ma, and Q.~Yang, ``Practical active learning
  with model selection for small data,'' in \emph{2021 20th IEEE International
  Conference on Machine Learning and Applications (ICMLA)}.\hskip 1em plus
  0.5em minus 0.4em\relax IEEE, 2021, pp. 1647--1653.

\bibitem{sener2017active}
O.~Sener and S.~Savarese, ``Active learning for convolutional neural networks:
  A core-set approach,'' \emph{arXiv preprint arXiv:1708.00489}, 2017.

\bibitem{nie2011survey}
C.~Nie and H.~Leung, ``A survey of combinatorial testing,'' \emph{ACM Computing
  Surveys (CSUR)}, vol.~43, no.~2, pp. 1--29, 2011.

\bibitem{22621}
\BIBentryALTinterwordspacing
D.~Kuhn, R.~Kacker, and Y.~Lei, ``\BIBforeignlanguage{en}{Combinatorial
  testing},'' 2012-06-25 2012. [Online]. Available:
  \url{https://tsapps.nist.gov/publication/get_pdf.cfm?pub_id=910001}
\BIBentrySTDinterwordspacing

\bibitem{kuhn2013combinatorial}
D.~R. Kuhn, I.~D. Mendoza, R.~N. Kacker, and Y.~Lei, ``Combinatorial coverage
  measurement concepts and applications,'' in \emph{2013 IEEE Sixth
  International Conference on Software Testing, Verification and Validation
  Workshops}.\hskip 1em plus 0.5em minus 0.4em\relax IEEE, 2013, pp. 352--361.

\bibitem{Dua:2019}
\BIBentryALTinterwordspacing
D.~Dua and C.~Graff, ``{UCI} machine learning repository,'' 2017. [Online].
  Available: \url{http://archive.ics.uci.edu/ml}
\BIBentrySTDinterwordspacing

\bibitem{harris2020array}
\BIBentryALTinterwordspacing
C.~R. Harris, K.~J. Millman, S.~J. van~der Walt, R.~Gommers, P.~Virtanen,
  D.~Cournapeau, E.~Wieser, J.~Taylor, S.~Berg, N.~J. Smith, R.~Kern, M.~Picus,
  S.~Hoyer, M.~H. van Kerkwijk, M.~Brett, A.~Haldane, J.~F. del R{\'{i}}o,
  M.~Wiebe, P.~Peterson, P.~G{\'{e}}rard-Marchant, K.~Sheppard, T.~Reddy,
  W.~Weckesser, H.~Abbasi, C.~Gohlke, and T.~E. Oliphant, ``Array programming
  with {NumPy},'' \emph{Nature}, vol. 585, no. 7825, pp. 357--362, Sep. 2020.
  [Online]. Available: \url{https://doi.org/10.1038/s41586-020-2649-2}
\BIBentrySTDinterwordspacing

\bibitem{scikit-learn}
F.~Pedregosa, G.~Varoquaux, A.~Gramfort, V.~Michel, B.~Thirion, O.~Grisel,
  M.~Blondel, P.~Prettenhofer, R.~Weiss, V.~Dubourg, J.~Vanderplas, A.~Passos,
  D.~Cournapeau, M.~Brucher, M.~Perrot, and E.~Duchesnay, ``Scikit-learn:
  Machine learning in {P}ython,'' \emph{Journal of Machine Learning Research},
  vol.~12, pp. 2825--2830, 2011.

\bibitem{Waskom2021}
\BIBentryALTinterwordspacing
M.~L. Waskom, ``seaborn: statistical data visualization,'' \emph{Journal of
  Open Source Software}, vol.~6, no.~60, p. 3021, 2021. [Online]. Available:
  \url{https://doi.org/10.21105/joss.03021}
\BIBentrySTDinterwordspacing

\end{thebibliography}

\end{document}